\documentclass{article}

\usepackage{microtype}
\usepackage{graphicx}
\usepackage{subfigure}
\usepackage{booktabs} 
\usepackage{amsmath}
\usepackage{amssymb}
\usepackage{mathtools}
\usepackage{xcolor}

\usepackage{hyperref}

\usepackage[accepted]{sysml2019}

\sysmltitlerunning{Efficient Winograd Convolution via Integer Arithmetic}

\begin{document}

\twocolumn[
\sysmltitle{Efficient Winograd Convolution via Integer Arithmetic}



\sysmlsetsymbol{equal}{*}

\begin{sysmlauthorlist}
\sysmlauthor{Lingchuan Meng}{arm}
\sysmlauthor{John Brothers}{arm}
\end{sysmlauthorlist}

\sysmlaffiliation{arm}{Arm Inc., San Jose, California, USA}

\sysmlcorrespondingauthor{Lingchuan Meng}{lingchuan.meng@arm.com}

\sysmlkeywords{Convolutional Neural Network, Power Efficiency, Winograd}

\vskip 0.3in

\begin{abstract}
Convolution is the core operation for many deep neural networks. The Winograd convolution algorithms have been shown to accelerate the widely-used small convolution sizes. Quantized neural networks can effectively reduce model sizes and improve inference speed, which leads to a wide variety of kernels and hardware accelerators that work with integer data. The state-of-the-art Winograd algorithms pose challenges for efficient implementation and execution by the integer kernels and accelerators. We introduce a new class of Winograd algorithms by extending the construction to the field of complex and propose optimizations that reduce the number of general multiplications. The new algorithm achieves an arithmetic complexity reduction of $3.13$x over the direct method and an efficiency gain up to $17.37\%$ over the rational algorithms. Furthermore, we design and implement an integer-based filter scaling scheme to effectively reduce the filter bit width by $30.77\%$ without any significant accuracy loss.

\end{abstract}
]



\printAffiliationsAndNotice{} 

\section{Introduction}
\label{sec:introduction}
Quantized convolutional neural networks (convnet) have been shown to work for inference with integer weights and activations \cite{krishnamoorthi2018quantization, warden2015eightbits}. By quantizing to 8-bit integers, model sizes can be reduced by a factor of four compared to the 32-bit floating-point models. Speedups of 2x-3x have been observed for quantized networks on CPUs compared to their floating-point counterparts. On hardware where optimized fixed-point capabilities are available, the speedup can reach up to 10x \cite{warden2017hvx}. Numerous efficient kernels with reduced-precision computation have achieved fast inference,  such as ARM CMSIS \cite{lai2018cmsis}, GEMMLOWP \cite{GLP}, Nvidia Tensor RT \cite{migcz2017tensor_rt}. Custom hardware \cite{sze2017_dnn, han2016eie, DLA} with reduced-precision has also been designed and built for fast inference.

Over $90\%$ of the computation in convnets during inference and training is in convolutions \cite{krizhevsky2012alexnet, szegedy15inceptionv3}. Different algorithmic methods have been devised to speed up this core operation. The methods include using the fast Fourier transform (FFT) \cite{mathieu2013fft, vasilache2014fbfft}, or the Winograd convolution algorithms \cite{winograd1980arithmetic, lavin2015winograd}. Particularly, the Winograd convolution has proved to work well for the typical small convolution sizes, such as $3\times 3$ in popular convnets, due to its arithmetic complexity reduction. However, the best-known Winograd algorithms for convnets are derived over the field of rationals $\mathbb{Q}$ \cite{lavin2015winograd} which exhibit undesirable overhead for full-precision implementation on custom inference accelerators with integer arithmetic.

These recent advances and limitations lead to the question: can we design efficient Winograd convolution algorithms and optimizations that use only integer arithmetic? This paper answers the question from both the algorithm perspective and implementation perspective with the main contributions as follows:
\begin{enumerate}
	\item We derive new complex Winograd convolution algorithms by extending the construction field from rationals to complex for convnet acceleration (Section \ref{subsec:complex_winograd}).
	\item We propose optimization techniques that effectively reduce the number of general multiplications in the complex algorithms, achieving an arithmetic reduction of $3.13$x over the direct method with the example. We also provide a quantitative analysis on the efficiency gain which ranges from  $15.93\%$ to $17.37\%$ over the best-known Winograd algorithms (Section \ref{subsec:complex_winograd}).
	\item We design and implement a hardware-friendly precision scaling scheme for Winograd-domain filters using integer arithmetic. The analysis shows a reduction of $30.77\%$ for filter bit width with very small static errors added (Section \ref{subsec:lossy_precision_scaling}).
	\item We evaluate modified quantized convnets where both the Winograd convolution and filter precision scaling are used. Compared to the reference models, there is no significant accuracy loss (Section \ref{subsec:efficiency_error_analysis}).
\end{enumerate}

\section{Related Work}
\label{sec:related_work}
The Winograd convolution algorithm was first used to accelerate convnets by \cite{lavin2015winograd}. The authors derived several small fixed-size algorithms over the field of rationals based on the minimal filtering algorithm proposed by Winograd \cite{winograd1980arithmetic}, which achieve arithmetic complexity reductions ranging from $2.25$x to $4$x for the popular filter sizes. 

Since then, many efforts have been made to improve the Winograd convolution. To address its numerical instability issue \cite{lavin2015winograd, budden2016deeptensor}, some mitigating techniques, such as using post-pass scaling of the convolution matrices, have been develped by \cite{vincent2017winograd} to enable larger tile sizes; \cite{barabasz2018winograd} found that the selection of good interpolation points depend on the values of the points and on symmetries between different points, meaning that sets of points with symmetric groups give better results. 

The Winograd algorithms decrease the sparsity of activations and filters when they are transformed into the Winograd domain. \cite{liu2018sparse_winograd} proposed two modifications to Winograd-based convnets to enable network pruning \cite{han2015deep_compression} to exploit sparsity: (1) moving the ReLU operation into the Winograd domain to increase the sparsity of activations, and (2) pruning the weights in the Winograd domain to exploit static weight sparsity. \cite{li2017sparse_winograd} introduced a Winograd layer in place of a standard convolution layer, which enables native pruning of Winograd coefficients and obtaining sparsity level beyond $90\%$. 

Efficient software implementations of the Winograd algorithms have also been developed. \cite{xygkis2018winograd} focused on the edge Internet of Things (IoT) devices where the computational resources are limited. \cite{jia2018winograd} proposed an algorithm for arbitrary-size N-dimensional Winograd-based convolution optimized for many-core CPUs, which achieves high hardware utilization through a series of optimizations. \cite{zlateski2018fft_winograd} proposed a performance model based on the Roofline mode \cite{williams2008roofline} to compare the Winograd approach and FFT-based approach and analyzed the conditions when one outperforms the other.

\section{Fast Algorithms}
\label{sec:fast_algorithms}
In this section, we review the fast algorithms for integer and complex multiplication and for short convolutions, namely the Karatsuba algorithm and the Winograd convolution algorithm. We analyze the best-known Winograd algorithms in the rational field $\mathbb{Q}$ and expose the challenges for integer accelerators to adopt the more efficient algorithms. We derive new convolution algorithms by extending to the field of complex $\mathbb{C}$ and analyze their arithmetic complexity and efficiency gains.

\subsection{Karatsuba Multiplication}
\label{subsec:karatsuba_multiplication}
The Karatsuba multiplication method is a classical divide-and-conquer algorithm that performs the multiplication of two $n$-digit numbers using at most $n^{log_2{3}} \approx n^{1.585}$ single-digit multiplications in general. 

Let $X$ and $Y$ be two $n$-digit numbers in some base $B$. The basic step of Karatsuba algorithm computes the product of $X$ and $Y$ using three multiplications and some additions and shifts. Let $m$ be any positive integer less than $n$, we write $X$ and $Y$ as
\begin{align*}
X &= x_0 + x_1B^m, \;Y = y_0 + y_1B^m,
\end{align*}
where $x_0$ and $y_0$ are the remainders of $X$ and $Y$ modulo $B^m$, and $x_1$ and $y_1$ are the quotients, respectively.
With this representation, The product of $X$ and $Y$ becomes
\begin{align*}
XY = x_0y_0 + (x_1y_0 + x_0y_1)B^{m} + x_1y_1B^{2m}.
\end{align*}

The Karatsuba algorithm computes the coefficient of $B^m$ as 
\begin{align*}
(x_1y_0 + x_0y_1) = (x_1 + x_0)(y_1 + y_0) - x_1y_1 - x_0y_0, 
\end{align*}
which reuses $x_1y_1$ and $x_0y_0$, leading to a multiplication of $X$ and $Y$ with three multiplications instead of four. 


The algorithm can also be used in complex multiplication, where the base $B$ is replaced with the imaginary unit $i$. The product of $X = x_0 +x_1\cdot i$ and $Y = y_0 + y_1\cdot i$ can be similarly computed with three multiplications as 
\begin{align*}
(\underbrace{x_0y_0}_\text{mul 1} - \underbrace{x_1y_1}_\text{mul 2}) + (\underbrace{(x_1 + x_0)(y_1 + y_0)}_\text{mul 3} - x_1y_1 - x_0y_0) \cdot i.
\end{align*} 

\subsection{Winograd Convolution}
\label{subsec:winograd_convolution}
The Winograd convolution algorithm generalizes the well-known method of the convolution theorem and fast Fourier transfrom (FFT) and outperforms it for short convolutions, as measured by the number of \textit{general} multiplications. 

Define a polynomial over a field $F$ as a mathematical expression
\begin{align*}
f(x) = f_nx^n + f_{n-1}x^{n-1} + \cdots + f_1x +f_0,
\end{align*}
where $x$ is symbolic and $f_0, \ldots,f_n$ are elements of the field $F$ known as the coefficients. Then convolutions can be formulated as polynomial products:
\begin{itemize}
\item Linear convolution can be written as $s(x) = g(x)d(x)$;
\item Cyclic convolution can be written as \\ $s(x) = g(x)d(x)\;  (\textnormal{mod}\;  x^n - 1)$. 
\end{itemize}

Fast convolution algorithms can be constructed with the Lagrange interpolation or the  Chinese remainder theorem (CRT) for polynomials. The Winograd convolution algorithm computes $s(x) = g(x)d(x)\;  (\textnormal{mod}\;  m(x))$, where $m(x)$, $g(x)$ and $d(x)$ are polynomials in $F$. The linear and cyclic convolutions can be trivially cast to this format. For example, setting $m(x) = x^n - 1$ yields the cyclic convolution. The algorithm breaks the problem into smaller pieces by factoring $m(x)$ into pairwise coprime polynomials $m^{(k)}(x)$ over a subfield of $F$ and constructs the solution using the CRT or interpolation. 

%

As in \cite{lavin2015winograd}, let $F(m, r)$ denote the computation of $m$ outputs with an $r$-tap FIR filter. $F(m, r)$ consumes $m + r - 1$ input values, the same number of general multiplications when computed with the Winograd algorithm. We express the algorithms in matrix form as
\begin{align*}
Y = A^{T}[(Gg)\odot(B^Td)],
\end{align*}
where $\odot$ represents element-wise multiplication, also known as the Hadamard product.

Higher dimensional algorithms $F(m \times n, r \times s)$ can be constructed by nesting the corresponding 1D algorithms $F(m, r)$ and $F(n, s)$ along each dimension. Particularly in convnets, square-shaped filters and activation patches are common, and a 2D algorithm $F(m\times m, r\times r)$ can be written as
\begin{align*}
Y = A^{T}[(GgG^T)\odot(B^TdB)]A, 
\end{align*}
whose arithmetic complexity reduction can be computed as 
\begin{align*}
\dfrac{m^2r^2}{(m+r-1)^2}.
\end{align*}
Therefore, the commonly-used algorithms such as $F(2\times 2, 3\times 3)$ and $F(4\times 4, 3\times 3)$ achieve reductions of $2.25$x and $4$x, respectively. 

In order to avoid additional general multiplications other than those in the Hadamard product $\odot$, \textit{good} interpolation points must be used in the derivation of Winograd algorithms \cite{blahut2010fast}. For $F(2, 3)$, $[0, 1, -1]$ are used to generate the auxiliary matrices that involve only additions, subtractions, and shifts by 1.

For $F(4\times 4, 3\times 3)$, the best-known algorithm is derived using the interpolation points at $[0, 1, -1, 2, -2]$. As introduced in  \cite{lavin2015winograd}, the filter transform matrix is 
\begin{align*}
G &= 
\begin{bmatrix*}[r]
\frac{1}{4} & 0 & 0\\
-\frac{1}{6} & -\frac{1}{6} & -\frac{1}{6}\\
-\frac{1}{6} & \frac{1}{6} & -\frac{1}{6}\\
\frac{1}{24} & \frac{1}{12} & \frac{1}{6}\\
\frac{1}{24} & -\frac{1}{12} & \frac{1}{6}\\
0 & 0 & 1
\end{bmatrix*}.
\end{align*}

$G$ and its transpose $G^T$ cause significant performance overhead for accelerators designed with integer arithmetic for quantized neural networks. Both matrices contain the large denominator of $24$ in its fractional values and have to be scaled up accordingly for full-precision integer arithmetic. This requires widening the spatial domain filter of $w$-bit by at least $\lceil log_2(24^2) \rceil = 10$ bits when it is transformed into the Winograd domain with $G$ and $G^T$, resulting a significant area increase for any custom integer multipliers that compute the element-wise multiplications in the Winograd domain.

To date, only the field of rationals $\mathbb{Q}$ has been used as the subfield of $F$ in the derivation of Winograd algorithms for neural network acceleration. Due to the undesirable numerical properties, most integer-based accelerators designed with Winograd convolution are limited to using $F(2\times 2, 3 \times 3)$ with only $2.25$x complexity reduction and its $1$D variants.

\subsection{Complex Winograd Convolution}
\label{subsec:complex_winograd}
We extend the subfield of $F$ from $\mathbb{Q}$ to the complex field $\mathbb{C}$ to derive new complex Winograd algorithms. This may seem counter-intuitive, as each multiplication in $\mathbb{C}$ takes four multiplications if implemented naively or three multiplications if the Karatsuba algorithm is used. Two key insights behind the complex Winograd are: (1) the symmetry of interpolation points and (2) the redundancy of information in complex arithmetic. The symmetry leads to the extension to the field of complex numbers. The redundancy leads to the optimization that exploits the complex conjugates. We will use $F(4\times 4, 3 \times 3)$ as an example throughout this section for derivation and optimization. 
\begin{align*}
B^T &= 
\begin{bmatrix*}[r]
1 & 0 & 0 & 0 & -1 & 0\\
0 & 1 & 1 & 1 & 1 & 0\\
0 & -1 & 1 & -1 & 1 & 0\\
0 & -i & -1 & i & 1 & 0\\
0 & i & -1 & -i & 1 & 0\\
0 & -1 & 0 & 0 & 0 & 1
\end{bmatrix*}, \\
G &= 
\begin{bmatrix*}[r]
1 & 0 & 0\\
\frac{1}{4} & \frac{1}{4} & \frac{1}{4}\\
\frac{1}{4} & -\frac{1}{4} & \frac{1}{4}\\
\frac{1}{4} & \frac{i}{4} & -\frac{1}{4}\\
\frac{1}{4} & -\frac{i}{4} & -\frac{1}{4}\\
0 & 0 & 1
\end{bmatrix*},
\\
A^T &=
\begin{bmatrix*}[r]
1 & 1 & 1 & 1 & 1 & 0\\
0 & 1 & -1 & i & -i & 0\\
0 & 1 & 1 & -1 & -1 & 0\\
0 & 1 & -1 & -i & i & 1\\
\end{bmatrix*}
\end{align*}

Recall that $F(4\times 4, 3 \times 3)$ requires five interpolation points. We replace the previously-known good points of $[0, 1, -1, 2, -2]$ in $\mathbb{Q}$ with $[0, 1, -1, i, -i]$ in $\mathbb{C}$, where $i$ is the imaginary unit. Using the same construction technique as in $\mathbb{Q}$, the new transform matrices for complex $F(4\times 4, 3 \times 3)$ can be generated as above.

By extending to and using the symmetric interpolation points in the complex plane, the magnitudes of elements in all three transform matrices have been reduced. $B^T$ and $A^T$ now only involve additions and subtractions. And the largest denominator in $G$ has been reduced from $24$ to $4$.

\subsection{Complexity Analysis}
\label{subsec:complexity_analysis}
This section analyzes the arithmetic complexity reduction of the new complex Winograd algorithm and shows how it reduces area and improves efficiency for integer arithmetic. 

First, we show an optimization technique that reduces the number of complex multiplications by exploiting the underlying complex conjugate pairs. The idea is simple: if we have calculated $x = a + bi$, then no additional  multiplication is needed for its complex conjugate $\overline{x} = a - bi$.
 
We use $B^TdB$ as an example. Let $d = [d_{i, j}]$ for $i, j \in [0, 1, \ldots, 5], d' = B^Td$ and $D = d'B$, then we have for $j = [0, 1, \ldots, 5]$
\begin{align*}
d'[0, j] &= d_{0, j} - d_{4, j}\\
d'[1, j] &=\sum^4_{k=1}d_{k, j}\\
d'[2, j] &= -d_{1, j} + d_{2, j} - d_{3, j} + d_{4, j}\\
d'[3, j] &= -d_{2, j} + d_{4, j} - (d_{1, j} - d_{3, j})i\\
d'[4, j] &= -d_{2, j} + d_{4, j} + (d_{1, j} - d_{3, j})i\\
d'[5, j] &= -d_{1, j} + d_{5, j}.
\end{align*}

The $[0, 1, 2, 5]$ rows contain only additions and subtractions among the integer input values. The $[3, 4]$ rows contain pairs of complex conjugates.

The same complex conjugate pattern can be found in the $[3, 4]$ columns in $D$ after $d'$ is right multiplied with $B$. Composing the patterns in rows of $B^T$ and columns of $B$, $D''$ contains the  complex conjugate pairs as
\begin{align*}
D = 
\begin{bmatrix*}[r]
D_{0, 0} & D_{0, 1} & D_{0, 2} & D_{0, 3} & \overline{D_{0, 3}} & D_{0, 5}\\
D_{1, 0} & D_{1, 1} & D_{1, 2} & D_{1, 3} & \overline{D_{1, 3}} & D_{1, 5}\\
D_{2, 0} & D_{2, 1} & D_{2, 2} & D_{2, 3} & \overline{D_{2, 3}} & D_{2, 5}\\
D_{3, 0} & D_{3, 1} & D_{3, 2} & D_{3, 3} & D_{3, 4} & D_{3, 5}\\
\overline{D_{3, 0}} & \overline{D_{3, 1}} & \overline{D_{3, 2}} & \overline{D_{3, 4}} & \overline{D_{3, 3}} & \overline{D_{3, 5}}\\
D_{5, 0} & D_{5, 1} & D_{5, 2} & D_{5, 3} & \overline{D_{5, 3}} & D_{5, 5}\\
\end{bmatrix*}
\end{align*}

That is, the $6\times 6$ transformed activation contains $10$ pairs of complex conjugates and the other $16$ values in $\mathbb{Q}$.

The same pattern can be found in the transformed filter $ W=GgG^T$ by noticing the rows $[3, 4]$ in $G$ are structurally the same as those in $B^T$, in terms of producing complex conjugate pairs. Therefore, we have
\begin{align*}
W = 
\begin{bmatrix*}[r]
W_{0, 0} & W_{0, 1} & W_{0, 2} & W_{0, 3} & \overline{W_{0, 3}} & W_{0, 5}\\
W_{1, 0} & W_{1, 1} & W_{1, 2} & W_{1, 3} & \overline{W_{1, 3}} & W_{1, 5}\\
W_{2, 0} & W_{2, 1} & W_{2, 2} & W_{2, 3} & \overline{W_{2, 3}} & W_{2, 5}\\
W_{3, 0} & W_{3, 1} & W_{3, 2} & W_{3, 3} & W_{3, 4} & W_{3, 5}\\
\overline{W_{3, 0}} & \overline{W_{3, 1}} & \overline{W_{3, 2}} & \overline{W_{3, 4}} & \overline{W_{3, 3}} & \overline{W_{3, 5}}\\
W_{5, 0} & W_{5, 1} &W_{5, 2} & W_{5, 3} & \overline{W_{5, 3}} & W_{5, 5}\\
\end{bmatrix*}
\end{align*}

Rewrite the 2D Winograd algorithm in matrix form:
\begin{align*}
Y &= A^{T}[(GgG^T)\odot(B^TdB)]A\\
   &= A^{T}[W\odot D]A
\end{align*}

Only the Hadamard product $W\odot D$ contains general multiplications. Furthermore, the complex values and their conjugates are at the matching positions in $D$ and $W$.  The $16$ pairs of rational elements, such as $\{D_{0, 0}, W_{0, 0}\}$, require $16$ general multiplications; The $20$ complex multiplications can be grouped into $10$ pairs of complex conjugate multiplications, such as $\{\{D_{0, 3}\cdot W_{0, 3}\}, \overline{\{D_{0, 3}}\cdot \overline{W_{0, 3}\}}\}$. Since $\overline{x}\cdot \overline{y} = \overline{x\cdot y}$, each set requires only one complex multiplication. Using the Karatsuba algorithm introduced in Section \ref{subsec:karatsuba_multiplication}, each complex multiplication takes 3 real multiplications. Therefore, the complex $F(4\times 4, 3\times 3)$ performs a total of $16 + 10\times 3 = 46$ general multiplications, leading to an arithmetic complexity reduction of $144/46 = 3.13$x as measured by the number of general multiplications. 

\textbf{Efficiency gain for hardware implementation.} Recall for the $F(4\times 4, 3\times 3)$ in $\mathbb{Q}$ with $4$x reduction, the bitwidth of Winograd-domain filters has to be widened by $\lceil log_2(24^2) \rceil = 10$ bits. With the $F(4\times 4, 3\times 3)$ in $\mathbb{C}$, the widening is reduced to $\lceil log_2(4^2) \rceil = 4$ bits. Given the typical bitwidth of spatial filters in quantized neural networks as $8$-bit, using the complex $F(4\times 4, 3\times 3)$ instead of its rational counterpart reduces the bitwidth by $1 - \dfrac{8 + 4}{8 + 10} = 33.33\%$ and achieves an efficiency gain of $\dfrac{3.13/12}{4.0/18}- 1 = 17.37\%$ with respect to the bitwidth. Comparing to the rational $F(2\times 2, 3\times 3)$, the efficiency gain is $\dfrac{3.13/(8 + 4)}{2.25/(8 + 2)}- 1 = 15.93\%$.

\textbf{Efficiency gain for software implementation.} Software speedup by CPU/GPU benefits from improved SIMD vectorization. The complex F(4x4, 3x3) reduces the bitwidth from 18 to 12, enabling int-16 SIMD instructions where available and extending an $n$-way vectorization to $2n$-way. 

Additional optimizations include:
\begin{itemize}
\item Keeping the Hadamard product in the Karatsuba format if the products are summed across multiple channels.
\item Skipping the calculations for the imaginary coefficients in the final results, as we know they will sum to $0$ because of the original computation of convolving two integer tensors $g$ and $d$.
\end{itemize} 

The optimization techniques and analysis developed in this section extend to the derivation of larger Winograd algorithms that require more good interpolations points in addition to $[0, 1, -1]$.

\section{Filter Precision Scaling}
\label{sec:filter_precision_scaling}
In this section, we propose an efficient precision scaling scheme for the Winograd-domain filters which further improves the efficiency of the Hadamard products without any significant accuracy loss. The scheme works in parallel with the complex Winograd algorithms introduced in Section \ref{subsec:complex_winograd}.

\subsection{Quantized Filters in Winograd Domain}
\label{subsec:quantized_filters}
For inference on mobile and edge devices, it has been shown that quantized neural network models can achieve comparable accuracies as the full-precision float-point (\textit{fp32}) models. Mainstream machine learning frameworks such as TensorFlow \cite{abadi2016tensorflow} have also developed their quantization flows that convert fp32 models to int8 models. As summarized in \cite{krishnamoorthi2018quantization}, typical quantization methods include: (1) uniform affine quantization, (2) uniform symmetric quantization, and (3) stochastic quantization.

\begin{figure}
	\includegraphics[width=0.5\textwidth]{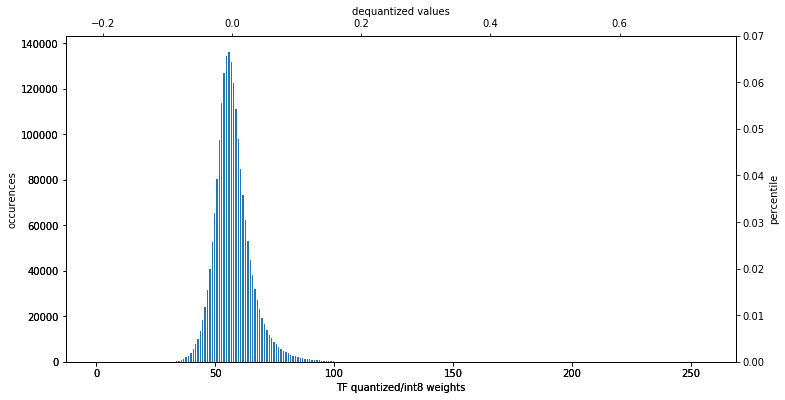}
	\vspace{-15pt}
	\caption{An example of TensorFlow-quantized weight distribution.}
	\label{fig:quant_weights}
\end{figure}

In this work, we assume the quantized filter weights are generated by the uniform affine quantization and represented in unsigned int8 (\textit{uint8}) together with a dynamic range. Figure \ref{fig:quant_weights} illustrates an example of the quantized weights extracted from a fully-connected layer. The bottom x-axis shows the uint8 weights ranging from $0$ to $255$; the top x-axis shows the dequantized fp32 weights. 

The spatial filters are transformed into the Winograd domain using the transform matrices. We use $F( 2\times 2, 3\times 3)$ in $\mathbb{Q}$ as an example. The filter transform matrix is $G'$
\begin{align*}
G' = 
\begin{bmatrix*}[r]
2 & 0 & 0\\
1& 1 & 1\\
1 & -1 & 1\\
0 & 0 & 2
\end{bmatrix*}
\end{align*}
which is produced by scaling up the original filter transform matrix $G$ \cite{lavin2015winograd} by a factor of $2$ element-wise for integer arithmetic.

Let $g = [g_{i, j}]$ for $i, j \in [0, 1, 2]$, then $G'g$ becomes
\begin{align*}
\begin{bmatrix*}[c]
2g_{0, 0}& 2g_{0, 1}& 2g_{0, 2}\\
g_{0, 0}\!+g_{1, 0}\!+g_{2, 0} & g_{0, 1}\!+g_{1, 1}\!+g_{2, 1} & g_{0, 2}\!+g_{1, 2}\!+g_{2, 2}\\
g_{0, 0}\!-g_{1, 0}\!+g_{2, 0} & g_{0, 1}\!-g_{1, 1}\!+g_{2, 1} & g_{0, 2}\!-g_{1, 2}\!+g_{2, 2}\\
2g_{2, 0} & 2g_{2, 1} & 2g_{2, 2}  
\end{bmatrix*}
\end{align*}
Denote the elements in $G'g$ as $p_i, i \in [0,1,\ldots,11]$ in the row-major order for a simpler representation, then $G'gG'^T$ becomes
\begin{align*}
\begin{bmatrix*}[c]
2p_0 & p_0 + p_1 + p_2 & p_0 - p_1 + p_2 & 2p_2 \\
2p_3 & p_3 + p_4 + p_5 & p_3 - p_4 + p_5 & 2p_5 \\
2p_6 & p_6 + p_7 + p_8 & p_6 - p_7 + p_8 & 2p_8 \\
2p_9 & p_9 + p_{10} + p_{11} & p_9 - p_{10} + p_{11}& 2p_{11}
\end{bmatrix*}.
\end{align*}

In order to adjust for the asymmetry introduced by the uniform affine quantization, the zero-point needs to be subtracted from the uint8 quantized weights, resulting in int9 weights ranging in $[-255:255]$. As a result, the worst-case magnitudes and bitwidths for each element in $G'gG'^T$ are
\begin{align*}
\begin{bmatrix*}[r]
1020 & 1530 & 1530 & 1020\\
1530 & 2295 & 2295 & 1530\\
1530 & 2295 & 2295 & 1530\\
1020 & 1530 & 1530 & 1020
\end{bmatrix*},
\begin{bmatrix*}[r]
11 & 12 & 12 & 11\\
12 & 13 & 13 & 12\\
12 & 13 & 13 & 12\\
11 & 12 & 12 & 11
\end{bmatrix*}.
\end{align*}

The same analysis can be applied to the activation transform $B^TdB$ whose results can be represented by 11-bit. In this paper, we focus on the filter precision scaling that can be preprocessed offline and incur no overhead at inference time.

\subsection{Filter Precision Scaling}
\label{subsec:lossy_precision_scaling}
Targeting quantized filters in the Winograd domain, we propose an efficient lossy precision scaling scheme using only integer arithmetic. We continue to use $F( 2\times 2, 3\times 3)$ as the running example.

The precision scaling is applied to the filters used to generate one output feature map (OFM). The scheme computes the minimum downscale factor at each X-Y location across all channels using the maximum magnitude. The scale factors are computed to put the transformed weights back into the int9 range which are then consumed by the multipliers. The downscaled Hadamard products are accumulated over all channels. Finally, the scaling is inverted before the final transforms $A^T$ and $A$.

Since the maximum magnitude in $G'gG'^T$ is $2295 = 9 \cdot 255$. The scale factors must cover the range of $\frac{1}{9}$ to $1$. For cost reasons, we implement the scaling by (1) multiplying with a $4$-bit number $n$, and (2) shifting right by a variable amount $p$.
That is, the scale factor is in the form of $\dfrac{n}{2^p}$. Recall that $F(2\times 2, 3\times 3)$ transforms a $3\times 3$ spatial filter to a $4\times 4$ Winograd-domain filter. Since we share the same scale factor at each X-Y location across all channels, $16$ scale factors are computed for a $4\times 4\times c$ Winograd-domain filter where $c$ is the number of channels.

Next we describe the steps to compute the scale factors.
\begin{enumerate}
	\item Transform all weights for the current OFM.
	\item Compute $16$ maximum magnitudes for the $4\times 4\times c$ Winograd-domain kernel.
	\item If the maximum magnitude for a given X-Y location $\leq 255$, set $n$ and $p$ to $0$, meaning no scaling applied.
	\item Otherwise, set the scale factor to the largest $4$-bit $n$ over $2^p$ by
	\begin{enumerate}
		\item Compute $x = \frac{255\cdot 128}{\text{max magnitude}}$
		\item Compute $y = \text{floor}(\text{log}_2(x))$
		\item $n = \text{floor}(\frac{x}{2^{y-4}})$
		\item $p = 7 - (y - 4)$.
	\end{enumerate}
\end{enumerate}
Note that $p$ ranges from $4$ to $7$ with an offset of $4$, therefore can be represented with $2$ bits. As a result, a total of $6$ bits are used to specify each scale factor, with the value $0$ meaning ``no scaling". The scaling factors are summarized in Table \ref{table:scaling} where the out-of-range and some duplicated scale factors are grayed out.

\begin{table}[t]
\caption{Downscaling factors for $F(2\times 2, 3\times 3)$ Winograd filters.}
\label{table:scaling}
\vskip 0.15in
\begin{center}
\begin{small}
\begin{sc}
\begin{tabular}{rcccr}
\toprule
N & P=0 & P=1 & P=2 & P=3 \\
\midrule
1   & \color{gray}{0.06250} & \color{gray}{0.03125} & \color{gray}{0.01563} & \color{gray}{0.00781} \\
2  & 0.12500 & \color{gray}{0.06250} & \color{gray}{0.03125} & \color{gray}{0.01563} \\
3  & 0.18750 & \color{gray}{0.09375} & \color{gray}{0.04688} & \color{gray}{0.02344} \\
4  & 0.25000 & 0.12500 & \color{gray}{0.06250} & \color{gray}{0.03125} \\
5  & 0.31250 & 0.15625 & \color{gray}{0.07813} & \color{gray}{0.03906} \\
6  & 0.37500 & 0.18750 & \color{gray}{0.09375} & \color{gray}{0.04688} \\
7  & 0.43750 & 0.21875 & 0.10938 & \color{gray}{0.05469} \\
8  & 0.50000 & 0.25000 & 0.12500 & \color{gray}{0.06250} \\
9  & 0.56250 & 0.28125 & 0.14063 & \color{gray}{0.07031} \\
10  & 0.62500 & 0.31250 & 0.15625 & \color{gray}{0.07813} \\
11  & 0.68750 & 0.34375 & 0.17188 & \color{gray}{0.08594} \\
12 & 0.75000 & 0.37500 & 0.18750 & \color{gray}{0.09375} \\
13  & 0.81250 & 0.40625 & 0.20313 & \color{gray}{0.10156} \\
14  & 0.87500 & 0.43750 & 0.21875 & 0.10938 \\
15  & 0.93750 & 0.46875 & 0.23438 & 0.11719\\
\bottomrule
\end{tabular}
\end{sc}
\end{small}
\end{center}
\vskip -0.1in
\end{table}

The reverse scaling before the final transforms is applied as a combination of 8-bit multiply and a right shift between 4 and 7 bits, which constitutes to a more precise approximation of the reciprocal of the corresponding scaling factor. During the application of $A^T$ and $A$, note that the original matrix $G$ has been scaled up by $2$ to $G'$, thus a right shift by $1$ must be taken after each final transform to cancel the scaling.

\subsection{Efficiency and Error Analysis}
\label{subsec:efficiency_error_analysis}
The transform and downscaling of the filters are performed before inference time, and Winograd-domain filters can be reused during inference. The downscaling step performs $O(n)$ comparisons, $4$-bit multiplications, and right shifts, where $n$ is the number of weights. The reverse scaling performs only $h\times w$ $8$-bit multiplications and $h\times w$ right shifts for an entire $h\times w \times c$ filter in the Winograd-domain, which reduces the amortized overhead effectively. 

Recall that subtracting the zero-point extends the range of quantized weights from uint8 to int9. In the example of $F(2\times 2, 3\times 3)$, applying the Winograd transforms further extends the required range to 13-bit as calculated in Section \ref{subsec:quantized_filters}. By using the proposed lossy filter precision scaling scheme, we reduce the range of Winograd weights back to 9-bit, leading to a filter bitwidth reduction of $30.77\%$.

The integer approximations of scaling factors introduce errors. We analyze the static errors here and measure the dynamic data-driven errors in terms of inference accuracy loss in Section \ref{subsec:experiments}. 

\begin{figure*}[!h]
	\includegraphics[width=\textwidth]{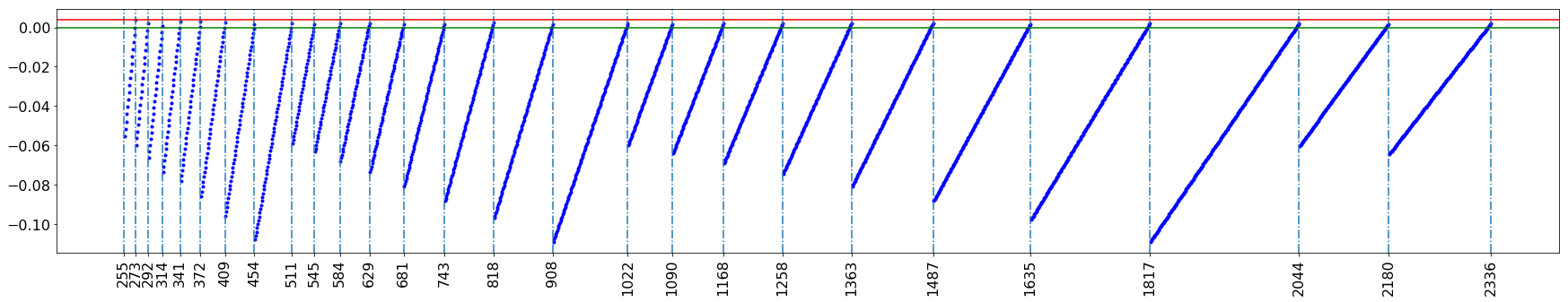}
	\vspace{-15pt}
	\caption{Relative errors within the applicable boundaries of each unique $\frac{n}{2^p}$ scaling factor.}
	\label{fig:range_error}
\end{figure*}

\begin{figure*}[!h]
	\includegraphics[width=\textwidth]{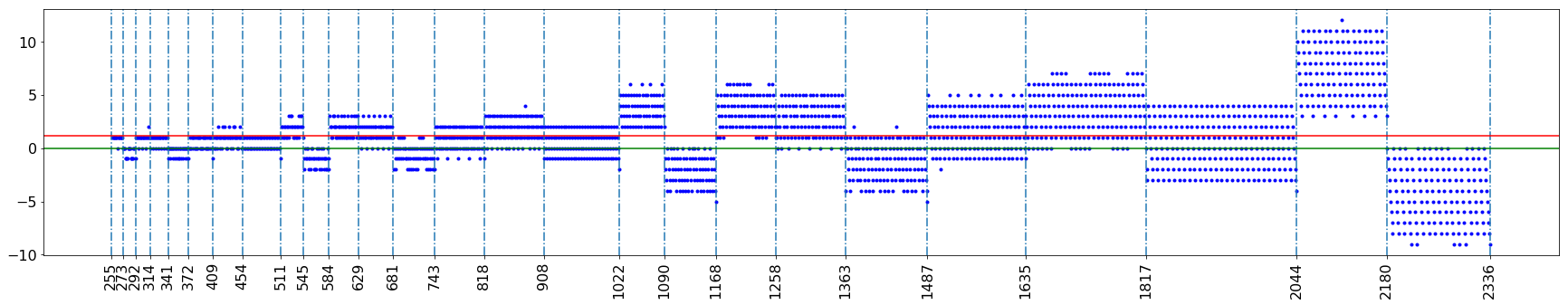}
	\vspace{-15pt}
	\caption{Numerical errors for scalable weights in the Winograd domain. The red line indicates the average error of $1.12$.}
	\label{fig:num_error}
\end{figure*}

\begin{figure*}[!h]
	\includegraphics[width=\textwidth]{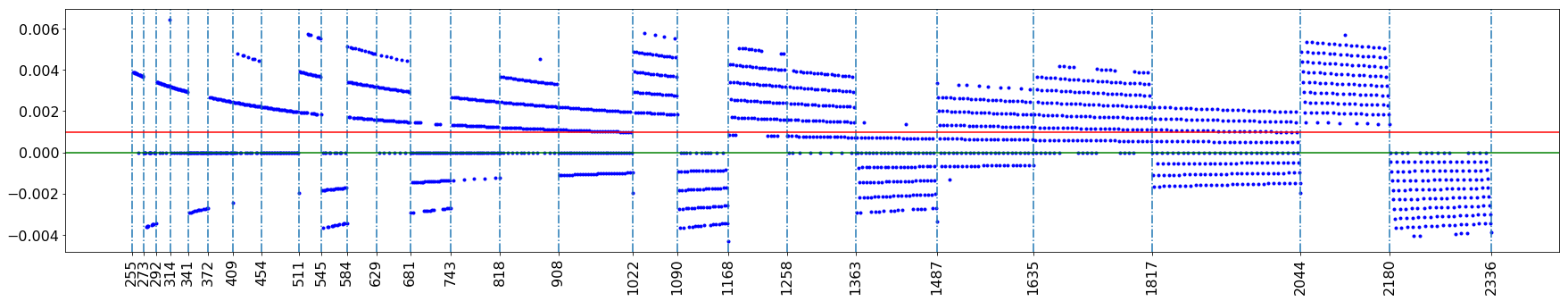}
	\vspace{-15pt}
	\caption{Proportional errors for scalable weights in the Winograd domain. The red line indicates the average error of $0.1\%$.}
	\label{fig:per_error}
\end{figure*}

For static scaling errors, Figure \ref{fig:range_error} uses the dashed vertical lines to show the applicable boundaries and the proportional scaling errors of each unique $\frac{n}{2^p}$ scaling factor. Figure \ref{fig:num_error} and \ref{fig:per_error} describe the numerical and proportional errors of all the scalable weights after being downscaled and then upscaled by the best integer-approximated scaling factors. The average numerical errors is $1.12$, and the average proportional error is $0.1\%$, indicating the filter precision scaling scheme introduces a small positive-biased error overall.


\subsection{Evaluation}
\label{subsec:experiments}
The filter precision scaling scheme is tested on the combination of popular convnet models of Inception V3 \cite{szegedy15inceptionv3} and ResNet V2 50, and a benchmark dataset ILSCVR-12 \cite{russakovsky14imagenet}. To produce the quantized models, we first obtain the pre-trained fp32 models published by TensorFlow-Slim \cite{silberman16tfslim}. Then we apply the standard quantization approach recommended by TensorFlow \cite{TFQ}. 

\begin{figure}[!t]
	\includegraphics[width=.45\textwidth]{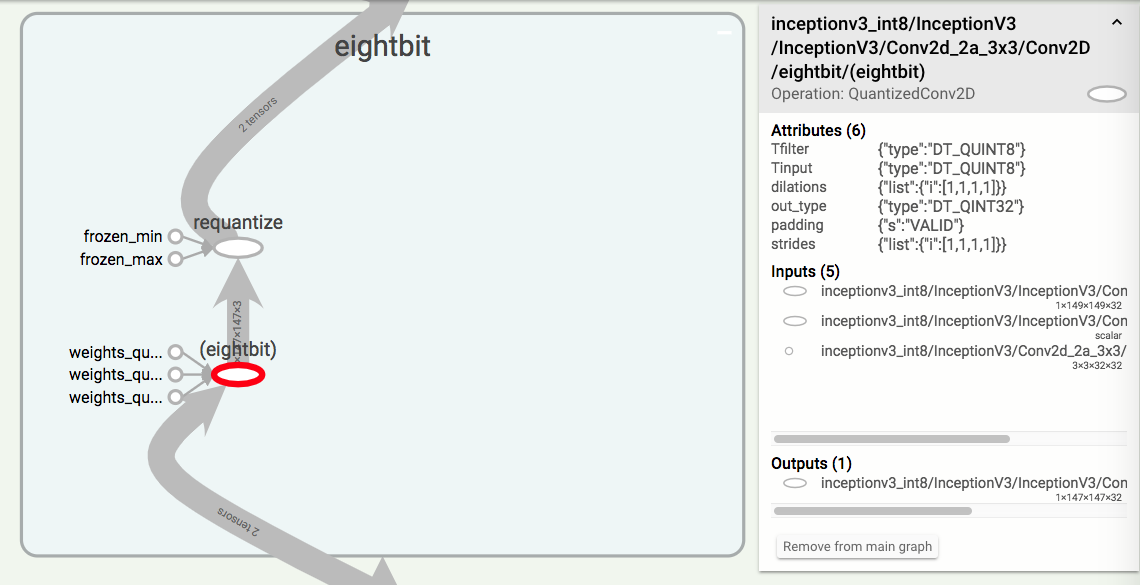}
	\vspace{-5pt}
	\caption{An example of QuantizedConv2D in Inception V3.}
	\label{fig:quant_conv2d}
\end{figure}

The quantization method replaces the fp32 \textit{Conv2D} nodes in the original pre-trained model with int8 \textit{QuantizedConv2D} nodes (usually followed by \textit{Requantize}), an example of which is illustrated by Figure \ref{fig:quant_conv2d} using the TensorBoard \cite{TB}. 

Our experiment captures a subset of QuantizedConv2D nodes in the quantized models where the filter height and width are both $3$ and the strides and dilations are both $[1, 1, 1, 1]$. Note that the nodes with non-unit strides or dilations or of $1$D shapes ($1\times3$ or $3\times1$) do not affect accuracy and are therefore skipped. The subsets of captured nodes, twelve for Inception V3 and sixteen for ResNet V2 50, are then edited dynamically using the Graph Editor library \cite{GE}. The editing takes place on two levels:
\begin{itemize}
	\item On the graph level, the subgraph that contains the captured nodes is duplicated within the same graph, such that the same image will be processed by both the reference subgraph and the Winograd and scaling-enabled counterpart.
	\item On the node level, each captured QuantizedConv2D node in the duplicated subgraph is replaced with a custom-built $F(2\times 2, 3\times 3)$ convolution scaled by the filter precision scaling method proposed in Section \ref{subsec:lossy_precision_scaling}.
\end{itemize}

\textbf{Inception V3}. The quantized model records a $73.91\%$ top-1 accuracy and a $90.97\%$ top-5 accuracy, the precision-scaled Winograd model achieves a $73.69\%$ top-1 accuracy $(\Delta = -0.22\%)$ and a $90.3\%$ top-5 accuracy $(\Delta = -0.67\%)$. 

\textbf{ResNet V2 50}. The precision-scaled Winograd model leads to a small loss of $0.13\%$ for top-1 accuracy ($73.34\% \rightarrow 73.21\%$) and the same top-5 accuracy $90.83\%$, compared to the quantized counterpart. 

Both experiments confirmed the proposed precision scaling scheme leads to very small accuracy loss for quantized models. Extensive experiments on more neural networks are planned as part of the future work.

\section{Conclusion}
\label{sec:Conclusion}
The Winograd convolution has proved its advantages for the small convolution sizes with the reduction in arithmetic complexity, but also poses challenges to efficient implementation by the emerging kernels and hardware accelerators with integer arithmetic. 

This paper is the first to extend the algorithm construction to the field of complex and derive the new complex algorithms for convnets. As an example, the complex $F(4\times 4, 3\times 3)$ achieves a complexity reduction of $3.13$x over the direct method and an efficiency gain in the range of $15.93\%$ to $17.37\%$ over the best-known rational Winograd algorithms with the hardware bitwidth is considered. The derivation method and optimization techniques developed in this paper extend to the construction of larger Winograd convolutions. This paper also answers the challenges from the implementation perspective. We proposed a fast integer-based precision scaling scheme for Winograd-domain filters. The scheme has been analyzed to show a significant reduction in filter bit width with very small static errors. Furthermore, we have shown the combination of Winograd convolution and the lossy scaling scheme can achieve good inference accuracy compared to the reference model without any significant loss. 

For future work, a quantitative impact analysis of the additional optimizations listed for complex Winograd convolution will be extended, and more experiments on the impact of precision scaling in a wider range of neural networks will be performed.


\section*{Acknowledgements}
The authors thank Jens Olson, Andrew Mundy, Rune Holm, Ian Bratt, Eric Kunze and Danny Loh for the insightful discussions and valuable feedback on this paper.


\bibliography{example_paper}
\bibliographystyle{sysml2019}


\end{document}